# Local and global approaches of affinity propagation clustering for large scale data*

Ding-yin XIA†, Fei WU†‡, Xu-qing ZHANG, Yue-ting ZHUANG
(*School of Computer Science and Technology, Zhejiang University, Hangzhou 310027, China*)
†E-mail: xiady@zju.edu.cn; wufei@zju.edu.cn

**Abstract:** Recently a new clustering algorithm called 'affinity propagation' (AP) has been proposed, which efficiently clustered sparsely related data by passing messages between data points. However, we want to cluster large scale data where the similarities are not sparse in many cases. This paper presents two variants of AP for grouping large scale data with a dense similarity matrix. The local approach is partition affinity propagation (PAP) and the global method is landmark affinity propagation (LAP). PAP passes messages in the subsets of data first and then merges them as the number of initial step of iterations; it can effectively reduce the number of iterations of clustering. LAP passes messages between the landmark data points first and then clusters non-landmark data points; it is a large global approximation method to speed up clustering. Experiments are conducted on many datasets, such as random data points, manifold subspaces, images of faces and Chinese calligraphy, and the results demonstrate that the two approaches are feasible and practicable.

## INTRODUCTION

Clustering is traditionally a fundamental problem in data mining. Clustering data based on the similarity value between data points is a key step in scientific data analysis and engineering. Usually we need to select some cluster centers to guarantee that the sum of squared errors between each data point and its potential cluster center is small during clustering.

Classical techniques for clustering, such as $k$-means clustering (MacQueen, 1967), partition the data into $k$ clusters and are very sensitive to the initial set of data centers, so they often need to be rerun many times in order to obtain a satisfactory result. Spectral clustering has its origin in spectral graph partition (Donath and Hoffman, 1973; Fiedler, 1973), and became a popular approach in high performance computing (Pothen *et al*., 1990). Support vector clustering (Ben-Hur *et al*., 2001) maps data points to a high-dimensional feature space by means of a Gaussian kernel, where the minimal enclosing sphere is searched. And the Markov cluster algorithm (Enright *et al*., 2002) is a fast and scalable unsupervised clustering algorithm for graphs based on the simulation of (stochastic) flow in graphs.

However, the abovementioned clustering approaches have not gained wide acceptance in practice due to their prohibitive cost and impracticality for real-world large scale data. Recently a new clustering approach named 'affinity propagation' (AP for short) (Frey and Dueck, 2007) has been devised to resolve these problems. The AP clustering method has been shown to be useful for many applications in face images, gene expressions and text summarization.

‡ Corresponding author
* Project supported by the National Natural Science Foundation of China (Nos. 60533090 and 60603096), the National Hi-Tech Research and Development Program (863) of China (No. 2006AA010107), the Key Technology R&D Program of China (No. 2006BAH02A13-4), the Program for Changjiang Scholars and Innovative Research Team in University of China (No. IRT0652), and the Cultivation Fund of the Key Scientific and Technical Innovation Project of MOE, China (No. 706033)

Unlike previous methods, AP simultaneously considers all data points as potential exemplars, and it recursively transmits real-valued messages along edges of the network until a good set of centers is generated. In particular, corresponding clusters gradually emerge in AP.

However, AP as well as most of the other clustering approaches aims to cluster data points with sparse relationship and is not suitable for a large dense similarity matrix, such as the Netflix dataset (Bell *et al.*, 2007). And the Netflix Prize seeks to fill in missing ratings and make the rating matrix dense. In this paper we present two methods for clustering the data for a dense similarity matrix. The results provide higher computational efficiency, greater stability and theoretical tractability. Partition affinity propagation (PAP) is an extension of AP which can reduce the number of iterations effectively and has the same accuracy as AP. This extension comes at the cost of making a uniform sampling assumption about the data. Landmark affinity propagation (LAP) is an approach for approximating a large global computation in clustering by a much smaller set of calculations, especially when the similarity matrix is dense. Most of the work in LAP focuses on a small subset of the large scale data, called 'landmark data points'.

The remainder of this paper is organized as follows. In Section 2 the AP clustering approach is summarized. In Section 3 we describe a perspective on the PAP method to reduce the number of iterations of clustering. In Section 4 we derive the LAP from a landmark version of clustering. In Section 5 experimental evaluations of random data points, manifold subspace data, face images and Chinese calligraphy show that our strategy generates efficient and effective clusters. Finally we conclude the paper and outline directions for future work in Section 6.

## AFFINITY PROPAGATION

AP (Frey and Dueck, 2006; 2007) takes as input a collection of real-valued similarities between data points, where the similarity $s(i, k)$ indicates how well the data point with index $k$ is suited to be the class center for data point $i$. When the goal is to minimize the squared error, each similarity is set to a negative Euclidean distance: for points $\boldsymbol{x}_i$ and $\boldsymbol{x}_k$, $s(i, k)=-\|\boldsymbol{x}_i-\boldsymbol{x}_k\|^2$.

Rather than requiring that the number of clusters be prespecified, AP takes as input a real number $s(k, k)$ for each data point $k$ so that data points with larger values of $s(k, k)$ are more likely to be chosen as class centers. These values are referred to as 'preferences'.

AP can be viewed as searching over valid configurations of the labels $c=\{c_1, c_2, \ldots, c_n\}$ to minimize the energy:

$$E(c) = -\sum_{i=1}^{N} s(i,\ c_i).$$

The process of AP can be viewed as a message communication process on a factor graph (Kschischang *et al.*, 2001). There are two kinds of messages exchanged between data points, i.e., 'responsibility' and 'availability'. The responsibility $r(i, k)$, sent from data point $i$ to candidate exemplar point $k$, reflects the accumulated evidence for how well-suited point $k$ is to serve as the exemplar for point $i$, taking into account other potential exemplars for point $i$. The availability $a(i, k)$, sent from candidate exemplar point $k$ to point $i$, reflects the accumulated evidence for how appropriate it would be for point $i$ to choose point $k$ as its exemplar, taking into account the support from other points that point $k$ should be an exemplar. The messages need only be exchanged between pairs of points with known similarities.

**Algorithm 1** Affinity Propagation

Input:

$s(i, k)$: the similarity of point $i$ to point $k$.

$p(j)$: the preferences array which indicates the preference that data point $j$ is chosen as a cluster center.

Output:

$idx(j)$: the index of the cluster center for data point $j$.

*dpsim*: the sum of the similarities of the data points to their cluster centers.

*expref*: the sum of the preferences of the identified cluster centers.

*netsim*: the net similarity (sum of the data point similarities and preferences).

Step 1: Initialize the availabilities $a(i, k)$ to zero:

$$a(i, k)=0. \tag{1}$$

Step 2: Update the responsibilities using the rule

$$r(i,k) \leftarrow s(i,k) - \max_{k' \text{ s.t. } k' \neq k} \{a(i,k') + s(i,k')\}. \tag{2}$$

Step 3: Update the availability using the rule

$$a(i,k) \leftarrow \min\Big\{0, r(k,k) + \sum_{i' \text{ s.t. } i' \neq i,k} \max\{0, r(i',k)\}\Big\}. \tag{3}$$

The self-availability is updated differently:

$$a(k, k) \leftarrow \sum_{i' \text{ s.t. } i' \neq k} \max\{0, r(i', k)\}. \quad (4)$$

Step 4: The message-passing procedure may be terminated after a fixed number of iterations, after changes in the messages fall below a threshold or after the local decisions stay constant for some number of iterations.

Availabilities and responsibilities can be combined to make the exemplar decisions. For point $i$, the value of $k$ that maximizes $a(i, k)+r(i, k)$ either identifies point $i$ as an exemplar if $k=i$ or identifies the data point that is the exemplar for point $i$. When updating the messages, numerical oscillations must be taken into consideration. As a result, each message is set to $\lambda$ times its value from the previous iteration plus $1-\lambda$ times its prescribed updated value. The $\lambda$ should be larger than or equal to 0.5 and less than 1. If $\lambda$ is very large, numerical oscillation may be avoided, but this is not guaranteed. Hence a maximal number of iterations are set to avoid infinite iteration in AP clustering.

## PARTITION AFFINITY PROPAGATION

The premise is that AP is a message-communication process between data points in a dense matrix. The time spent is in direct ratio to the number of iterations. During an iteration of AP, each element $r(i, k)$ of the responsibility matrix must be calculated once and each calculation must be applied to $N-1$ elements, where $N$ is the size of the input similarity matrix, according to Eq.(2). And each element of the availability matrix can be calculated in the same way.

The algorithm we propose reduces the time spent by decreasing the number of iterations by decomposing the original similarity matrix into sub-matrices (Guha *et al.*, 2001). The procedure of PAP is stated below:

**Algorithm 2** Partition Affinity Propagation

Input: The same as that in Algorithm 1.

Output: The same as that in Algorithm 1.

Step 1: Partition the matrix $\boldsymbol{S}_{N\times N}$ into $k$ parts as an average. At each iteration, $1<k<N/(4C)$, where $C$ is the maximal number of clusters prospected. So the sub-matrices $\boldsymbol{S}_{11}, \boldsymbol{S}_{22}, \ldots, \boldsymbol{S}_{kk}$ are all square matrices. The sizes of $\boldsymbol{S}_{11}, \boldsymbol{S}_{22}, \ldots, \boldsymbol{S}_{(k-1)(k-1)}$ are all $m=\lfloor N/k \rfloor$, which is an integer less than or equal to $N/k$, and the size of $\boldsymbol{S}_{kk}$ is $N-m(k-1)$.

$$\boldsymbol{S} = \begin{pmatrix} \boldsymbol{S}_{11} & \boldsymbol{S}_{12} & \cdots & \boldsymbol{S}_{1k} \\ \boldsymbol{S}_{21} & \boldsymbol{S}_{22} & \cdots & \boldsymbol{S}_{2k} \\ \vdots & \vdots & & \vdots \\ \boldsymbol{S}_{k1} & \boldsymbol{S}_{k2} & \cdots & \boldsymbol{S}_{kk} \end{pmatrix}.$$

Step 2: Use sub-matrices $\boldsymbol{S}_{11}, \boldsymbol{S}_{22}, \ldots, \boldsymbol{S}_{kk}$ as the input of AP, and run them respectively, then we obtain $k$ availability matrices:

$$\boldsymbol{A}_{11}, \boldsymbol{A}_{22}, \ldots, \boldsymbol{A}_{kk}.$$

Step 3: Combine $\boldsymbol{A}_{11}, \boldsymbol{A}_{22}, \ldots, \boldsymbol{A}_{kk}$ using the following rule to form the availability matrix of the whole dataset, setting the other parts of matrix $\boldsymbol{A}'$ to zeros:

$$\boldsymbol{A}' = \begin{pmatrix} \boldsymbol{A}_{11} & \boldsymbol{0} & \cdots & \boldsymbol{0} \\ \boldsymbol{0} & \boldsymbol{A}_{22} & \cdots & \boldsymbol{0} \\ \vdots & \vdots & & \vdots \\ \boldsymbol{0} & \boldsymbol{0} & \cdots & \boldsymbol{A}_{kk} \end{pmatrix}.$$

Step 4: Run AP using $\boldsymbol{A}'$ as the initial availability matrix until convergence.

PAP algorithm works as follows. First the large scale data points are divided into several divisions (Step 1) and the messages are passed in each division (Step 2). After that, each data point will have several cluster center candidates within its division. Then $\boldsymbol{A}_{11}$, $\boldsymbol{A}_{22}$, ..., $\boldsymbol{A}_{kk}$ are put together to communicate with the results obtained from other divisions (Steps 3 and 4). Since the availabilities $a(i, k)$ are initialized to zeros, PAP can reduce the number of iterations effectively.

In the final step, grouping results would be obtained in less time (i.e., fewer iterations) in most cases than by putting all the data points together at the very beginning. Furthermore, as in the final step grouping is done at all data points, the grouping result is similar to or even better than when putting all the data points together at the very beginning.

As the data are randomly distributed, it is reasonable to suppose that the time used for seeking the maximum value of an array is in direct ratio to its size. In Step 2, as the size of $\boldsymbol{S}_{ii}$ is about $1/k^2$ of $\boldsymbol{S}_{N\times N}$, each iteration of $\boldsymbol{S}_{ii}$ spends about $1/k^2$ of the time of $\boldsymbol{S}_{N\times N}$. With the size of the similarity matrix decreasing, the average number of iterations taken for convergence also decreases. The time complexity of AP is $O(n^2)$. When the size of the input similarity matrix $\boldsymbol{S}$ and the number of partition parts $k$ increase by some scale, the time spent on the sub-matrices can be ignored. So the total time spent in PAP is obviously shortened.

## LANDMARK AFFINITY PROPAGATION

LAP is a technique for approximating a large global computation in clustering by a much smaller set of calculations. Most of the work focuses on a small subset of the data, called 'landmark data points' (de Silva and Tenenbaum, 2003; 2004). Fig.1 shows the dynamics of LAP applied to 500 2D data points with 100 landmark points, using the negative squared error as the similarity.

Suppose we are given the similarity of $N$ data points denoted by matrix $\boldsymbol{S}_{N\times N}$, which is the same as the input of AP.

**Algorithm 3** Landmark Affinity Propagation

Input: The same as that in Algorithm 1.

Output: The same as that in Algorithm 1.

Step 1: Given a very large dataset $P$, which contains $n$ ($n$>10 000) data. And select a small dataset $M$ which contains $k$ ($k$<$n$) data by random sampling as landmark points. Then applying AP to the dataset $M$, we will obtain $c$ centers and $c$ classes.

Step 2: Calculate the maximum distance of the points to their class centers in each class, *max_distance*($k$). Then embed the remaining points ($P-M$) into the $c$ classes as follows:

When point $i$ in ($P-M$) is assigned to class $c_j$, subject to

$$distance(i, c_j)<max_distance(c_j);$$
$$distance(i, c_j)=\min\{distance(i, c_1), distance(i, c_2), \ldots, distance(i, c_n)\}.$$

But there are still $n$ points that do not belong to any classes.

Step 3: If $m$<$m_0$ ($m_0$ is the max-size of the data that can be calculated by AP clustering on a computer), apply AP; if $m$>$m_0$, apply LAP recursively. After that, merge the class centers from Step 1 and Step 3.

Step 4: Refine the final set of centers and classes, recalculate the similarity and obtain the best final results.

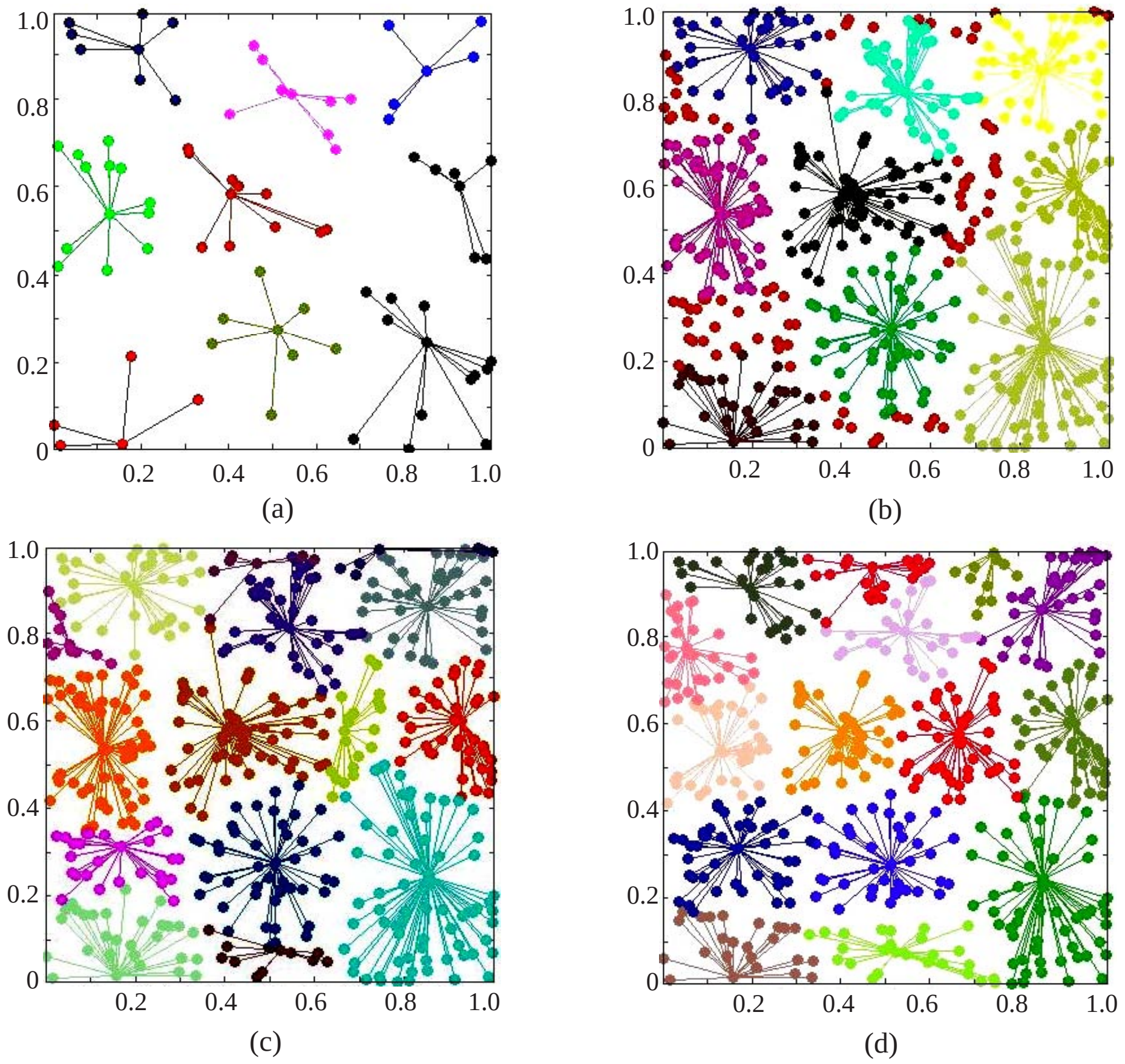

**Fig.1 Applying LAP to cluster 500 randomly selected 2D data points with 100 landmark points. (a)~(d) represent Steps 1~4 in Algorithm 3**

The simplest and most efficient approach for selecting the landmark point set $M$ is uniform random sampling. However, this approach may result in overrepresentation of larger clusters in the dataset and omission of smaller ones. So we would prefer a new approach for selecting landmarks in manifold learning based on Least Absolute value Subset Selection Operator (LASSO) regression (Silva *et al.*, 2005), which shows that non-uniform sampling of landmarks for manifold learning can give parsimonious approximations using very few landmarks. We will pursue research in this direction in our future studies.

## EXPERIMENTAL RESULTS

In this section we present a set of experiments to verify the effectiveness and efficiency of our proposed algorithms for data clustering. For comparison, we also report the experimental results from AP.

Performances of the algorithms were evaluated on four datasets:

(1) Two-dimensional data points by random selection.

(2) Several manifold subspace data points: Swiss Roll, Punctured Sphere, Gaussian, Corner Planes, Twin Peaks, by rolling up a randomly sampled plane into a spiral.

(3) Cohn-Kanade facial expression database (Kanade *et al.*, 2000). We sampled 62 points on each face contour such as eyebrows, eyes, nose, mouth and cheek, and considered the median of points sampled from the nose as the origin.

(4) Chinese calligraphy database. We selected 10729 images from the database in (Zhuang *et al.*, 2004), which includes 1496 Chinese characters, each with a different style.

Use the negative of squared Euclidian distance as the similarity between data points, and set $\lambda$ to be 0.5 in all the experiments. We utilized the number of iterations and the total time used for one run on AP, PAP and LAP as the evaluation metrics for clustering speed (by adjusting the input preference appropriately). All experiments were conducted on an Intel Quad-Core Xeon E5320 1.86 GHz processor with 16 GB memory, and the time was reported in seconds.

### Two-dimensional datasets

We tested 2D data point sets of size 100, 1000, 2000 and 3000 ten times in each group, respectively. Both AP and PAP were run on each group for comparison.

Fig.2a shows the average number of iterations of AP and the average number of iterations in Step 4 of PAP with $k$=4. Fig.2b shows the average time spent by AP and PAP. We can see that when the size of the dataset reaches a certain scale, the speed-up improvement is obvious.

We tested a set of 1000 2D data points with 100, 200, 300, 400 and 500 landmark points 10 times, respectively. Both AP and LAP were run on each group. Test results (Table 1) showed that the performances were not significantly degraded when LAP was used.

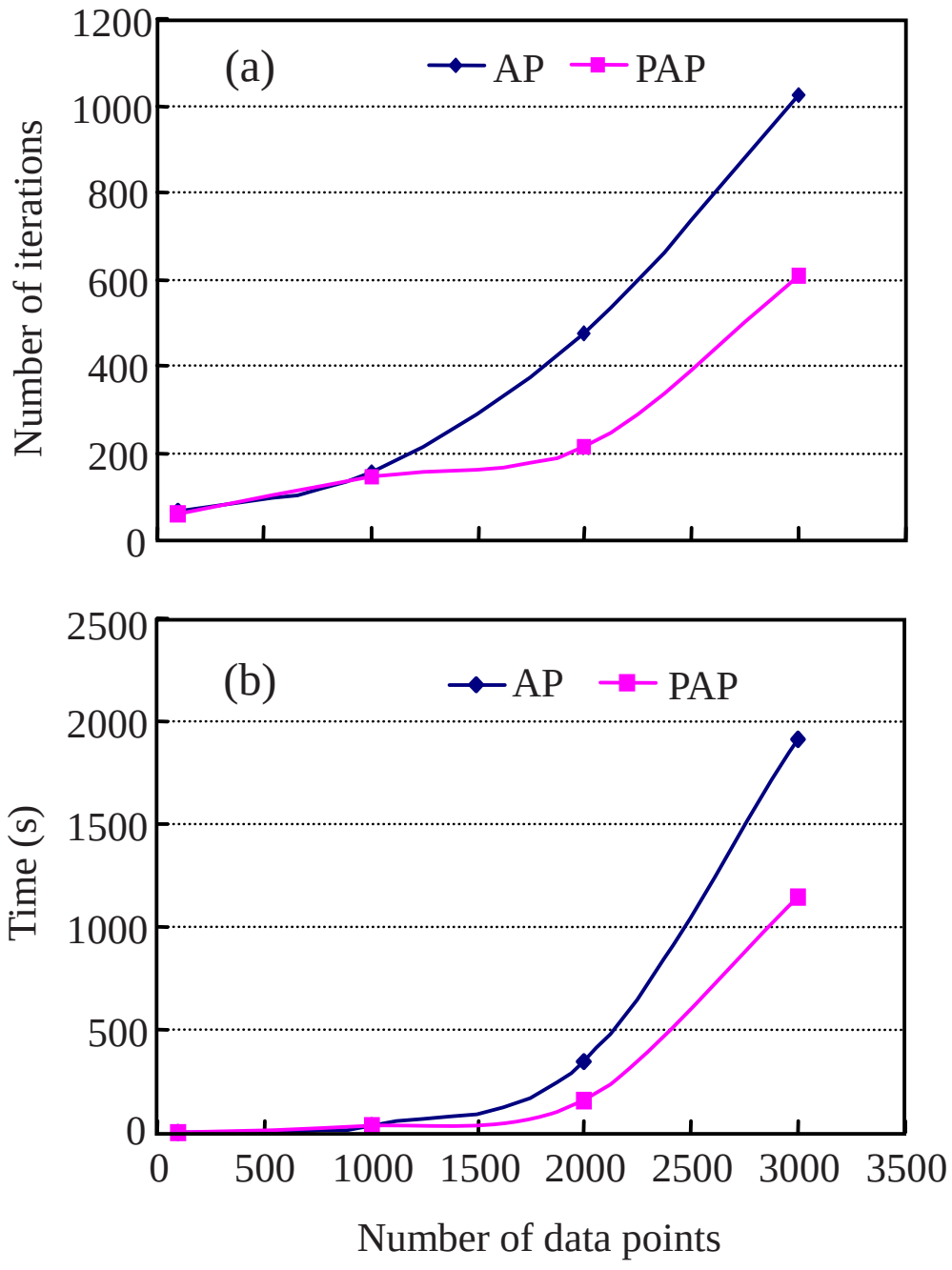

**Fig.2 Number of iterations (a) and computation time (b) of AP and PAP when $k$=4**

**Table 1 Statistics of running AP and LAP on 1000 randomly generated 2D data points**

| Algorithm | Number of landmark points | Time (s) | Accuracy (%) |
|---|---|---|---|
| AP | | 35.1506 | |
| LAP | 500 | 8.1697 | 81.90 |
| | 400 | 4.3074 | 69.40 |
| | 300 | 1.9303 | 57.90 |
| | 200 | 0.9365 | 45.40 |
| | 100 | 0.5034 | 29.60 |

**Manifold subspace datasets**

We tested Swiss Roll, Punctured Sphere, Gaussian, Corner Planes, and Twin Peaks datasets of 2000 3D data points generated by MANI (Wittman, 2005).

Fig.3 shows the cluster results of applying AP and PAP ($k$=4) to the Swiss Roll, Punctured Sphere, Gaussian, Corner Planes, and Twin Peaks datasets. It can be seen from Fig.3 that PAP maintains the original clustering accuracy, sometimes it even improves the results (Fig.3a).

**Cohn-Kanade AU-coded facial expression database test**

We selected 2021 images from the database, including images of 25 people, each with a different facial expression. Since the Cohn-Kanade facial expression database is manually labeled, we can calculate the error rates, including the 'false association rate' and the 'true association rate'. The 'true association rate' is the fraction of pairs of images from the same true category that were correctly placed in the same learned category, and the 'false association rate' is the fraction of pairs of images from different true categories that were erroneously placed in the same learned category. Table 2 gives the test results, which indicate that the speed-up obtained from PAP is dependent on $k$. If $k$ is well chosen, the average number of iterations in Step 4 of Algorithm 2 is only 10% of that in Algorithm 1. How to choose $k$ to gain the best speed-up is one of our further research tasks.

Fig.4 shows some of the clustering results obtained from running PAP with $k$=4. An image belonging to the same cluster was put in the same document. To represent the face images as visual content, a 62-dimensional feature (such as eyebrows, eyes, nose, mouth and cheek) was extracted from 2021 face images, and we considered the median of points sampled from the nose as the origin. Then we used the negative of the squared Euclidian distance as the similarity between images. It is stated as follows: for points $\boldsymbol{x}_i$ and $\boldsymbol{x}_k$, $s(i, k)=-\|\boldsymbol{x}_i-\boldsymbol{x}_k\|^2$.

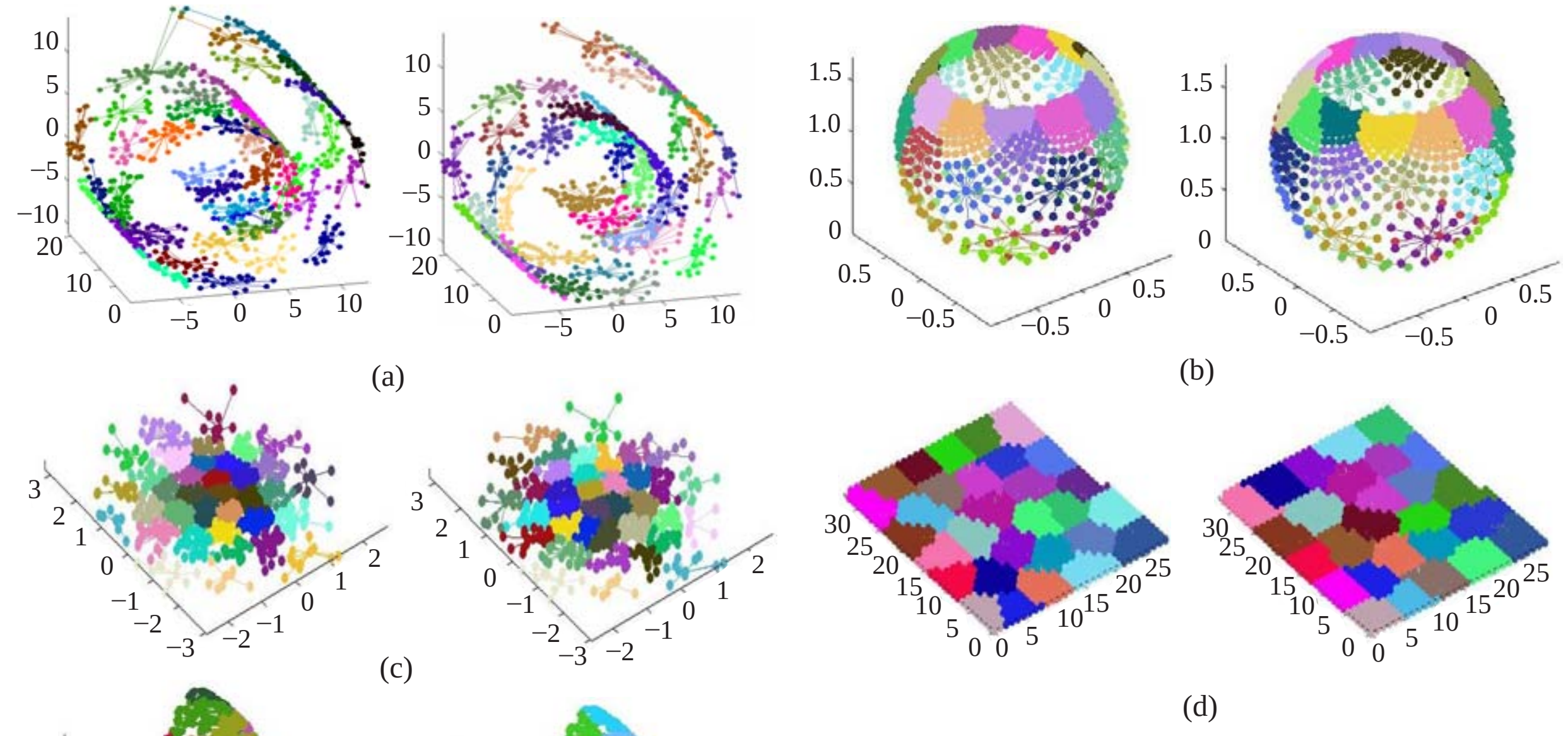

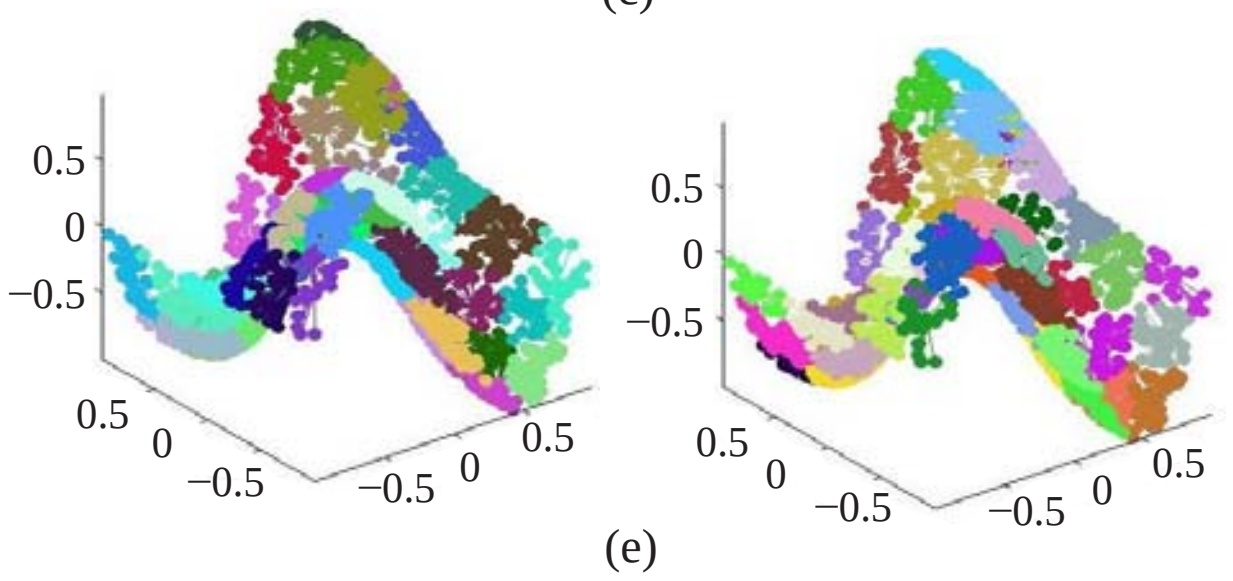

**Fig.3 Results of running AP and PAP ($k$=4) on several simple manifold datasets. (a) Swiss Roll dataset; (b) Punctured Sphere dataset; (c) Gaussian dataset; (d) Corner Planes dataset; (e) Twin Peaks dataset**

**Table 2 Statistics of running AP and PAP on 2021 images from the Cohn-Kanade facial expression database**

| Algorithm | $k$ | Number of classes | Number of iterations | Time (s) | Net similarity | True association rate (%) | False association rate (%) |
|---|---|---|---|---|---|---|---|
| AP | | 86 | 786 | 638 | −5.12 | 61.76 | 4.56 |
| PAP | 2 | 88 | 78 | 109 | −4.96 | 64.47 | 3.86 |
| | 4 | 87 | 97 | 151 | −4.92 | 65.61 | 3.57 |
| | 8 | 90 | 102 | 107 | −4.93 | 65.27 | 3.67 |
| | 16 | 85 | 109 | 128 | −5.18 | 60.84 | 4.81 |

### Chinese calligraphy database test

We selected 10729 images from the database, including 1496 Chinese characters, each with a different style. To depict each Chinese calligraphic character, a set of contour points (in general a feature of more than 150 dimensions) was extracted from each Chinese calligraphic character image. The total similarity of two character images is defined as follows [see Eqs.(2)~(5) in (Zhuang *et al.*, 2004)]:

$$TMC=\sum_{i=1}^{n}(PMC_i+\alpha\,\|\,p_i-corresp(p_i)\,\|^2), \quad (5)$$

where $\|p_i-corresp(p_i)\|^2$ is the Euclidean distance between point $p_i$ and the approximate corresponding point $corresp(p_i)$ in a candidate calligraphic character, and $PMC_i$ is the minimum point matching cost for $p_i$.

Fig.5 shows some of the clustering results after running AP on 100 Chinese calligraphy characters of 10 classes. Fig.6 shows some of the intermediate results after running Steps 1 and 2 of LAP and some of the final clustering results of LAP with 100 landmark points. The results show that LAP has a very good clustering performance.

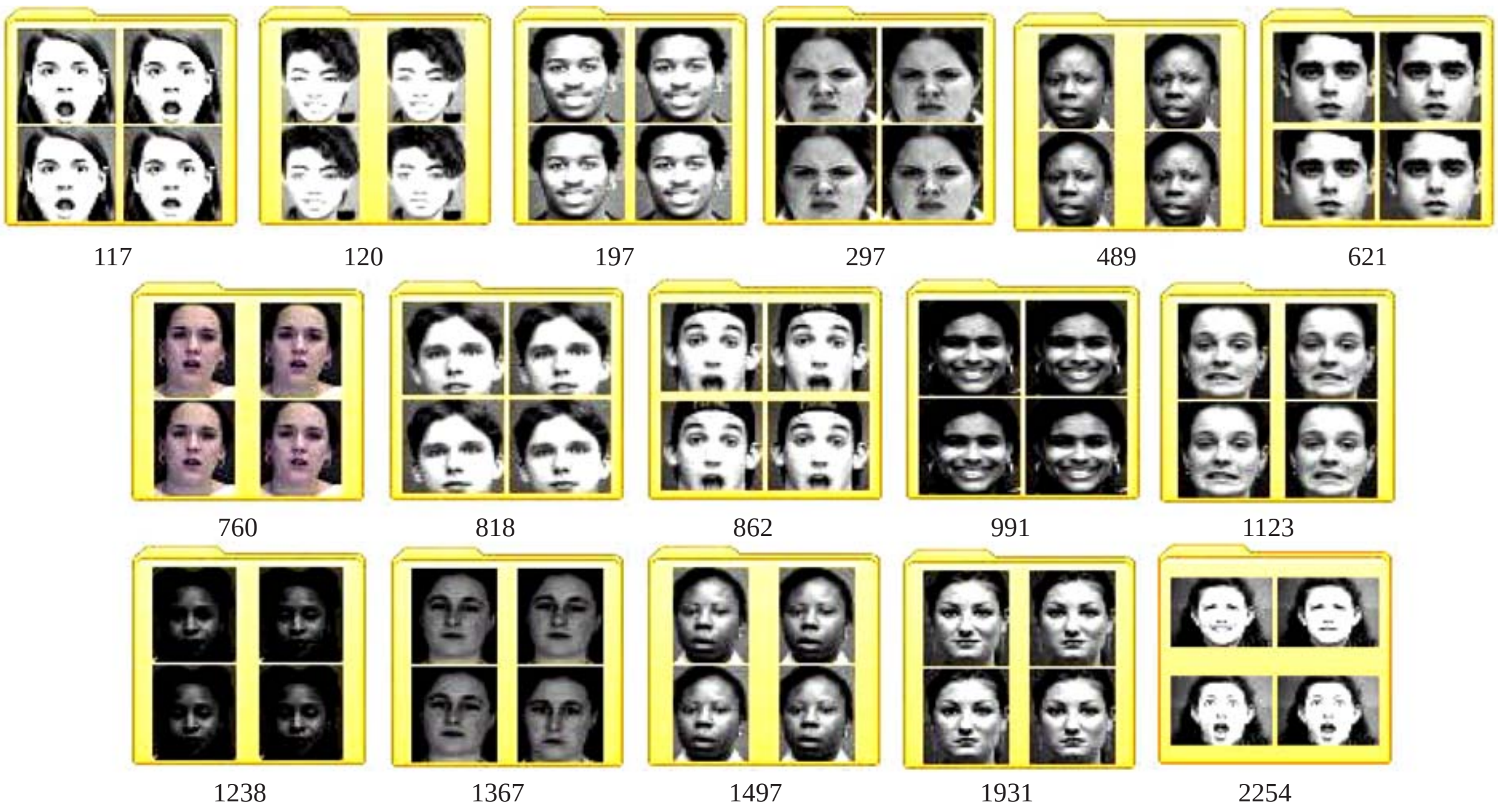

**Fig.4 Some of the clustering results from running PAP with $k$=4. The images come from the Cohn-Kanade AU-coded facial expression database. Each group of face pictures represents one class**

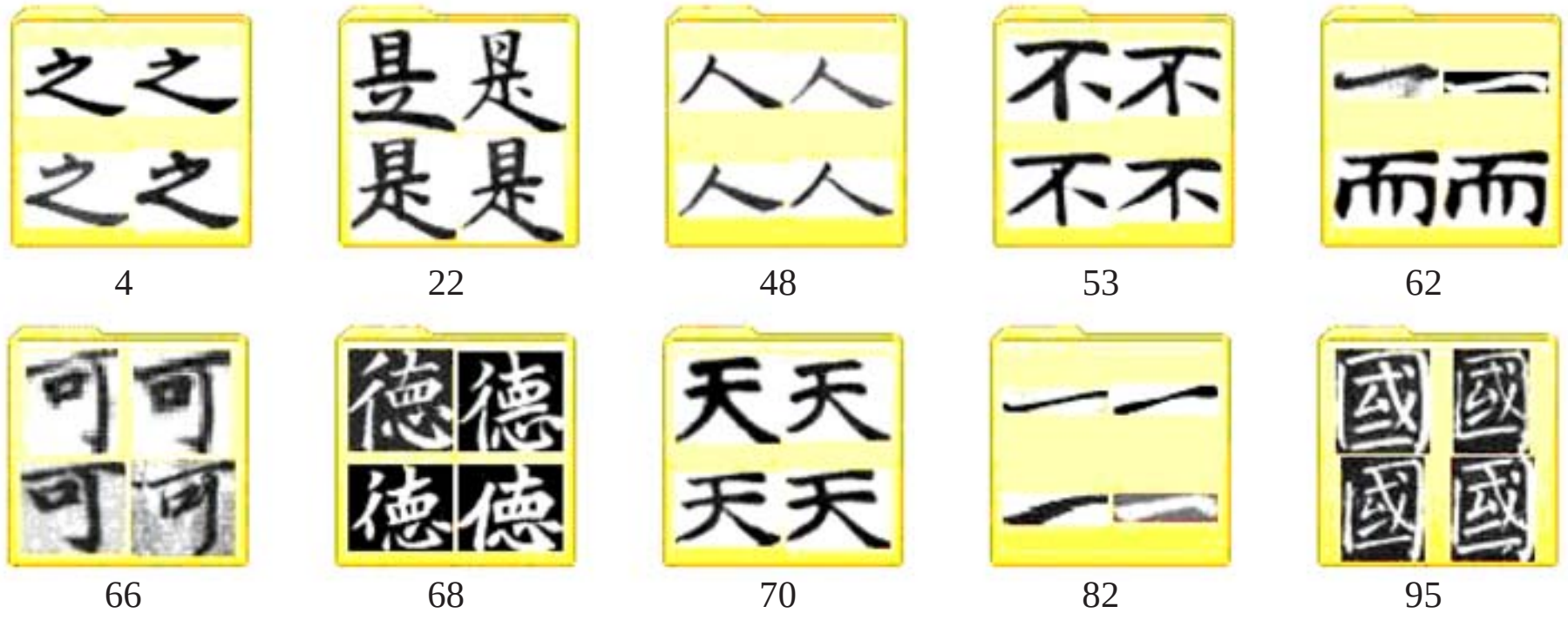

**Fig.5 Some of the clustering results after running AP on 100 Chinese calligraphy characters of 10 classes**

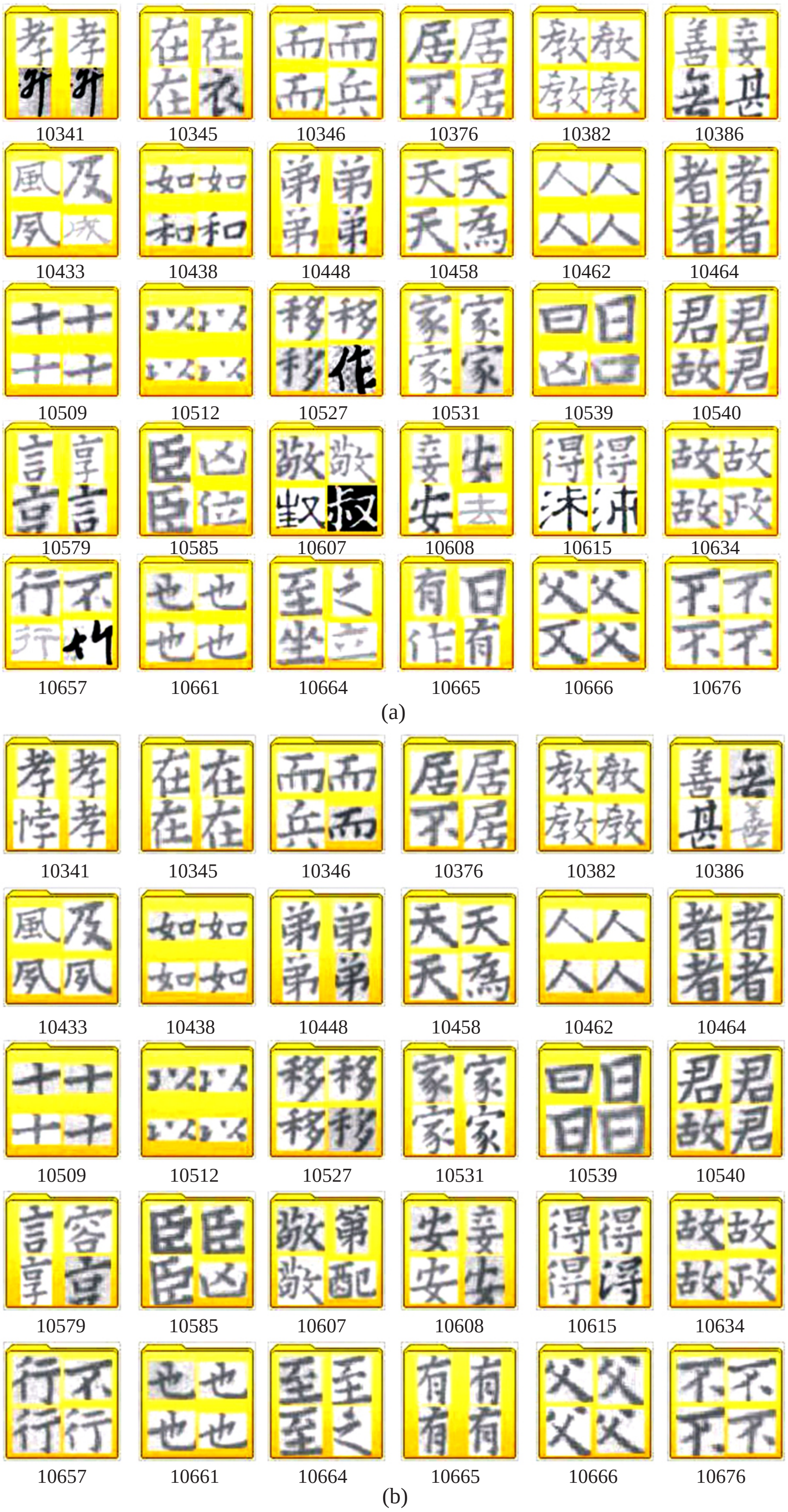

**Fig.6 (a) Some of the intermediate results after running Steps 1 and 2 of LAP on Chinese calligraphy characters; (b) Some of the final clustering results after running LAP on 10729 Chinese calligraphy characters with 100 landmark data points, where the similarity matrix is dense**

## CONCLUSION AND FUTURE WORK

To accelerate the AP algorithm for clustering a dataset with a dense relationship while maintaining its accuracy, an optimal algorithm, PAP, is proposed; the LAP is also presented for a large global approximation computation in clustering using a much smaller set of calculations. PAP and LAP speed up AP by passing messages in a subset of a large scale similarity matrix. Experiments were carried out on four datasets and the results demonstrated the effectiveness and efficiency of the proposed algorithms.

To further improve PAP, we will evaluate the best value of $k$ in PAP in the future and do several runs of LAP where the output exemplars can be used to better initialize Step 1 of LAP. This might be a good way to choose the landmark points which can give better approximations.